\def\year{2021}\relax
\documentclass[letterpaper]{article} 
\usepackage{aaai21}  
\usepackage{times}  
\usepackage{helvet} 
\usepackage{courier}  
\usepackage[hyphens]{url}  
\usepackage{graphicx} 
\usepackage{natbib}  
\usepackage{caption} 
\usepackage{amsmath}
\usepackage{xcolor}

\title{Using Human-Guided Causal Knowledge for More Generalized Robot Task Planning}
\author {
    Semir Tatlidil,\textsuperscript{\rm 1} 
    Yanqi Liu,\textsuperscript{\rm 2} 
    Emily Sheetz,\textsuperscript{\rm 3} 
    R. Iris Bahar,\textsuperscript{\rm 2} 
    Steven Sloman,\textsuperscript{\rm 1} \\
}
\affiliations {
    \textsuperscript{\rm 1} Dept.\/of Cognitive, Linguistics, and Psychological Sciences, Brown University, Providence, RI  02912 \\
    \textsuperscript{\rm 2} Dept.\/of Computer Science, Brown University, Providence, RI  02912 \\
    \textsuperscript{\rm 3} Dept.\/of Electrical Engineering and Computer Science, Robotics Institute, Univ.\/ of Michigan, Ann Arbor, MI 48109 \\    
    kemal\_semir\_tatlidil@brown.edu, yanqi\_liu@brown.edu, esheetz@umich.edu, iris\_bahar@brown.edu, steven\_sloman@brown.edu
}

\begin{document}

\maketitle

\begin{abstract}
A major challenge in research involving artificial intelligence (AI) is the development of algorithms that can find solutions to problems that can generalize to different environments and tasks. Unlike AI, humans are adept at finding solutions that can transfer.  We hypothesize this is because their solutions are informed by causal models. We propose to use human-guided causal knowledge to help robots find solutions that can generalize to a new environment. We develop and test the feasibility of a language interface that na{\"i}ve participants can use to communicate these causal models to a planner. We find preliminary evidence that participants are able to use our interface and generate causal models that achieve near-generalization. We outline an experiment aimed at testing far-generalization using our interface and describe our longer terms goals for these causal models.
\end{abstract}

\noindent

\section{Introduction}
One of the most impressive feats of human intelligence is our ability to \textit{transfer} learned knowledge to solve a problem in a novel environment  \citep{shepard1987toward}. Despite their overall success, state-of-the-art machine learning methods struggle with this \citep{edmonds2020theory, nair2019causal}. \par

What makes this problem challenging is determining an appropriate level of granularity for state representations. If an agent represents the states of its environment at a very detailed level, then it will likely fail to apply the solution that it learned in a new environment because even small changes in the environment will appear as new states. For example, consider the task of putting together an object that can emit light. In the training environment, we learn that connecting wax and a wick and then lighting the wick will emit light. If we try to apply this knowledge in a new environment where wax is replaced with a container filled with oil, we can only do this if we know that wax and oil share some critical feature that makes them interact with the wick in a similar way to emit light. In other words, we need to construct an abstract model of the states in our environment. \par

The importance of constructing abstract state representations for human-like intelligence has been emphasized by AI researchers \citep{KONIDARIS20191}. Previous works has shown that abstract state representations can be learned automatically from simulations using reinforcement learning (RL)  \citep{pmlr-v48-abel16} or generative adversarial nets \citep{kurutach2018modelensemble}. Studies focusing on transfer learning show that abstract representations lead to more efficient solutions in novel tasks \citep{asadi2020learning, Kokel_Manoharan_Natarajan_Ravindran_Tadepalli_2021}. However, a major drawback of these methods is that they may require many trials in order to learn a good abstract state representation, and the learned representations may not be interpretable by humans. \par

In contrast to AI research, studies with people suggest that they are able to construct abstract representations with limited experience and are able to generalize this knowledge across tasks \citep{holyoak2012analogy}. What makes people so efficient in constructing abstract representations? One line of evidence suggests that it is people's tendency to think about events in causal terms. Studies looking at how people create categories, which can be seen as a method of state abstraction by aggregating instances, show that people are heavily influenced by causal information in category formation \citep{Marsh2009, WALDMANN200627}. Moreover, they tend to employ these causal categories to solve problems of a similar nature in new environments \cite{LIEN200087}. In complex domains where a multitude of features can be used for categorization (such as categorizing different types of fish), evidence indicates that novices are more likely to categorize by considering \textit{surface level features} such as perceptual similarity.  On the other hand, experts are more likely to categorize based on \textit{causal information} \citep{Rottman2012, Shafto2003}. Overall, these studies provide strong evidence that humans tend to create abstract categories using causal information and use such information in a variety of (simple or complex) tasks. \par

The central role of causality in achieving human-like intelligence has been recognized by AI researchers particularly with respect to learning abstract, generalizable knowledge \citep{scholkopf2021toward, ZHU2020310}.  For instance, \citet{edmonds2020theory} investigated whether learning causal information in an escape room task would lead to better performance in a new environment. They showed that while a number of RL agents showed little to no transfer of knowledge across environments, an agent that learned abstract causal rules was able to transfer this knowledge to the new environment and learn the new solution much quicker. Moreover, the behavior of the causal agent was similar to the behavior of human subjects. This supports both the importance of causality in human thinking and how causal knowledge aids in generalizing knowledge across environments. In a different task that involved learning the relationship between many electrical switches and light bulbs, \citet{nair2019causal} have found that an agent that learns causal rules is better able to use this knowledge to solve tasks in new environments compared to model-free RL agents that do not learn the causal rules. Overall, the evidence strongly suggests that understanding the causal mechanisms of a task is critical for generalizing performance to new environments both in human agents and in AI. \par

Our goal is to develop a human-robot interaction system where causal knowledge communicated by a human is used by the robot to complete specific tasks and apply this knowledge to novel tasks. We propose to use causal information communicated to the robot by a human in the form of \textit {causal graphical models} \citep{sloman2005causal}. These causal models will communicate which features of the environment are relevant to the task, giving us an appropriate state abstraction. To the extent that a new task fits into the same causal model, this knowledge can be reused. 

In this paper, we present the first step of this work where we develop a user interface that allows people to communicate causal models of how various objects work to produce light. We conducted a preliminary online experiment to test if na{\"i}ve users could use our interface to express their causal models and to see if their causal models would achieve near-generalization.

\section{Assembling objects using causal models}

Our approach can be described in three steps. First, participants use an interface to express the causal model of how an object works to produce light. The users are required to express a causal graph in terms of what function each object part performs that causes light to be emitted, instead of describing the object parts themselves. Second, we interpret the user-generated causal graph as a reward function for a high-level symbolic planner. The planner generates a sequence of actions to assemble the object. Finally, the high-level symbolic planner will be connected to a simulator that converts the plan into a sequence of motor commands and visually demonstrates the execution of the task. In our experiment, we have four light-producing objects: a candle, a kerosene lamp, a desk lamp, and a flashlight. A candle and a kerosene lamp are considered to be from the same category of "light production method" as they both burn fuel to produce light. A desk lamp and a flashlight are considered to be from the same category as they both use electricity to produce light. \par 
Our definition of a successful causal model is a causal model that can be used by the planner to generate an accurate plan to assemble the new object, regardless of the specific labels chosen by the participants to represent the functions of the object parts. If the causal model of an object can be used to generate a plan for another object from the same category, we consider this a successful \textit{near-generalization}. If it can be used to generate a plan for an object from a different category, then it is a successful \textit{far-generalization} as shown in 
Fig.\ref{fig:table}. In this experiment, we only tested near-generalization.

\begin{table}[h!]
    \centering
    \begin{tabular}{|p{0.25\linewidth}|p{0.25\linewidth}|p{0.25\linewidth}|}
    \hline
        {\small Training object} & {\small Near-generalization} & {\small Far-generalization} \\
         \hline
          \includegraphics[width=0.7\linewidth]{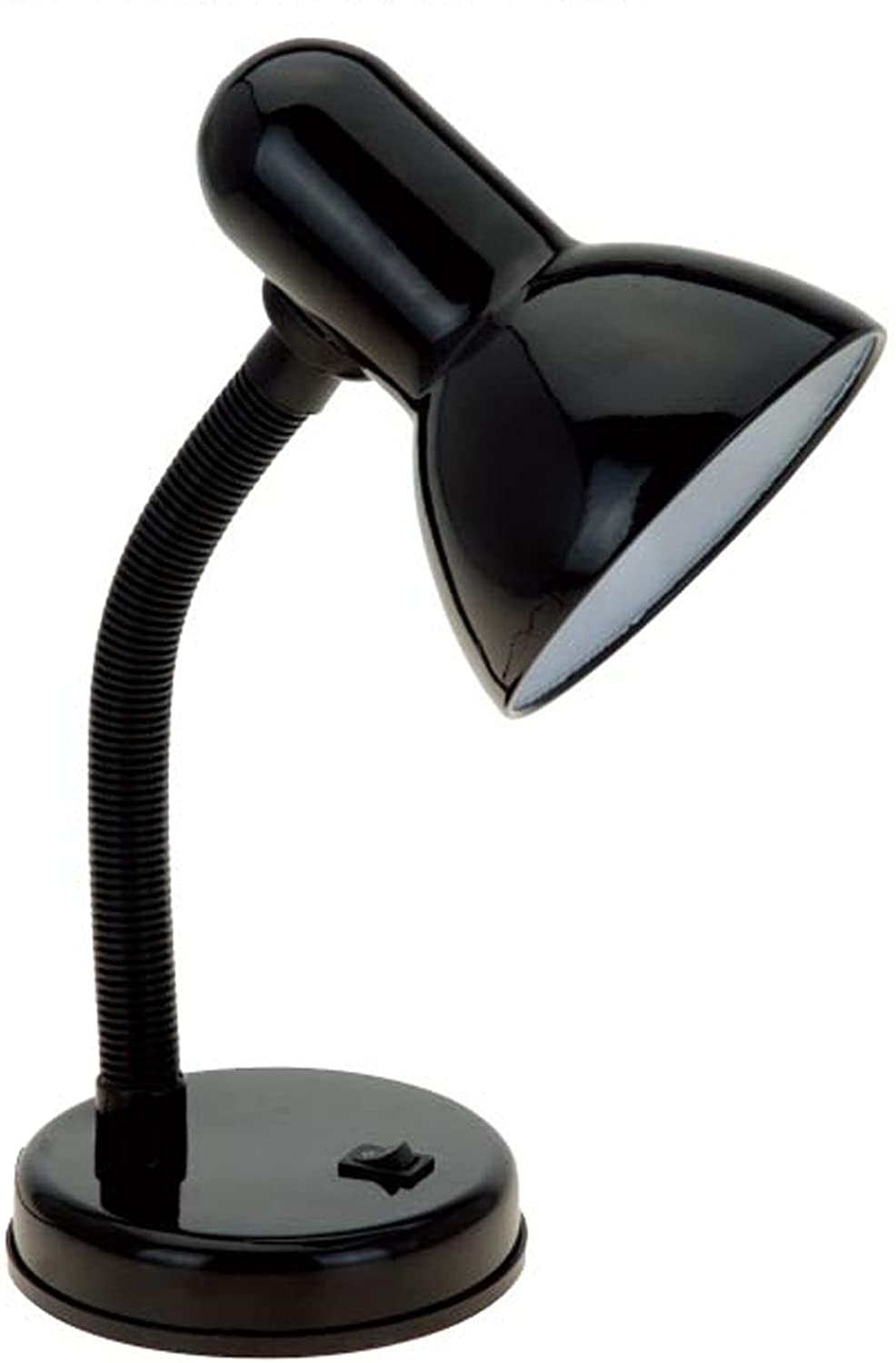}& \includegraphics[width=0.95\linewidth]{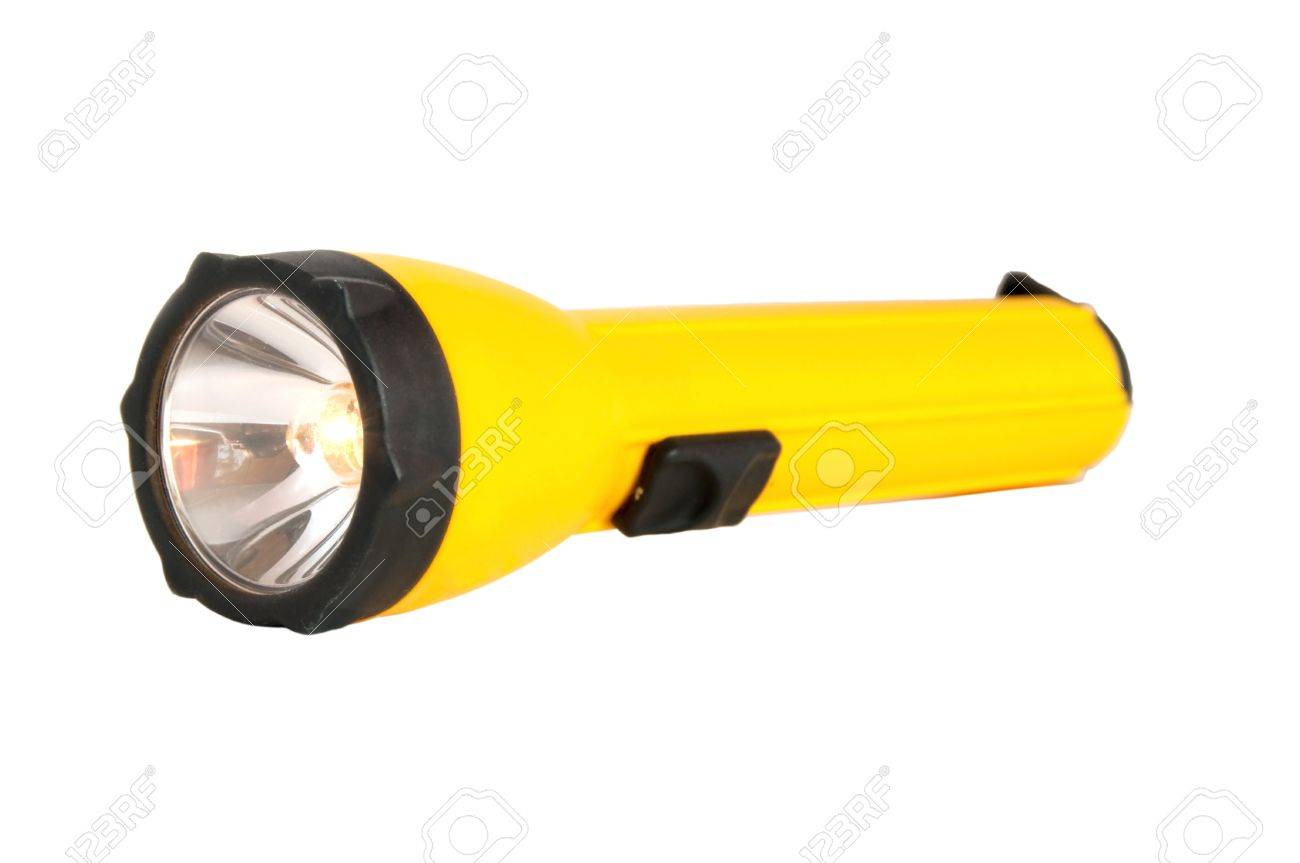} &
         \includegraphics[width=.95\linewidth]{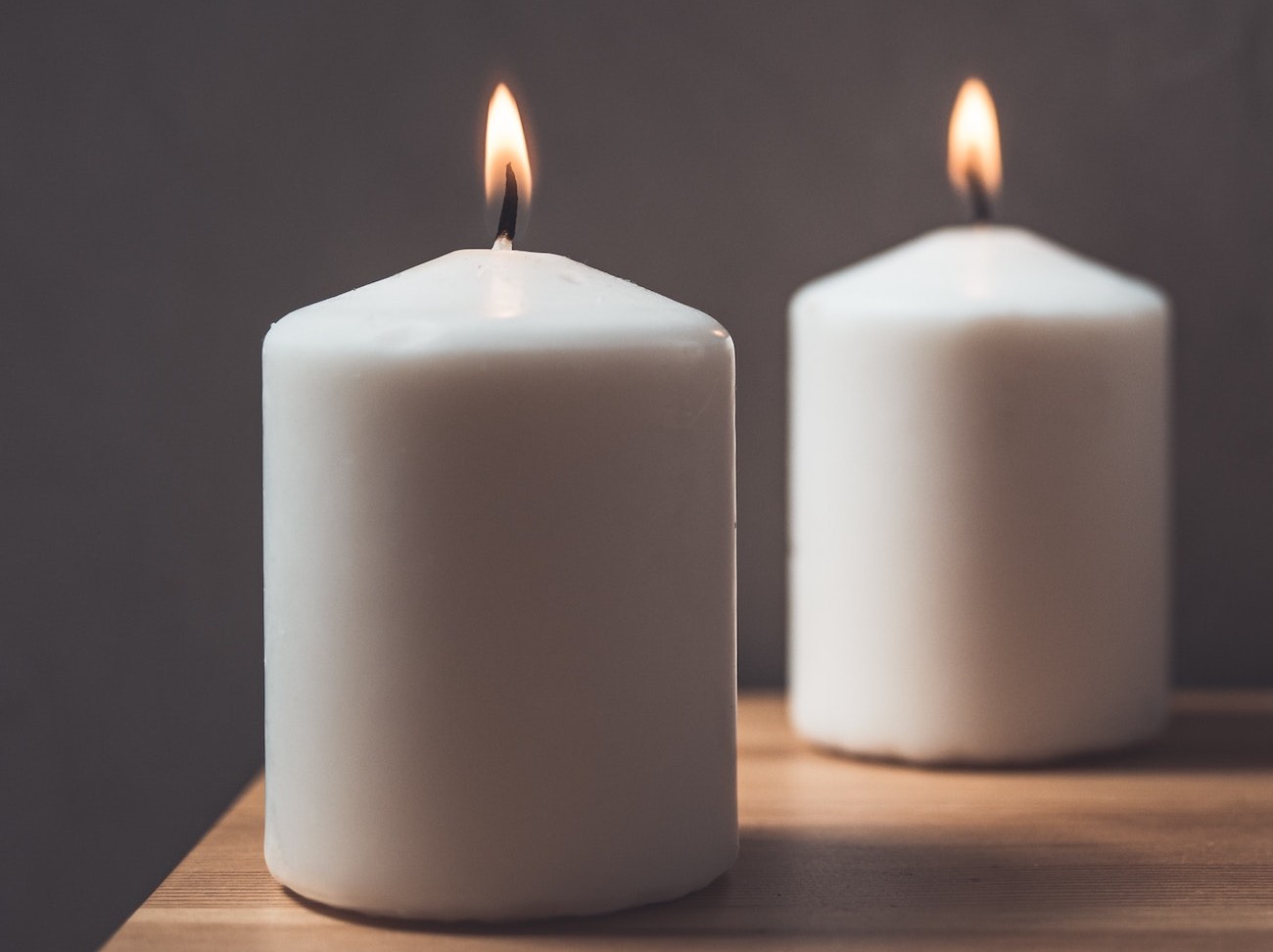} \\
         {\small Desk lamp} & {\small Flashlight} & {\small Candle} \\
         \hline
    \end{tabular}
    \captionof{figure}{ Near- and far-generalization}
    \label{fig:table}
\end{table}

\section{Near-generalization experiment}
\begin{figure}
    \centering
    \includegraphics[width=\linewidth]{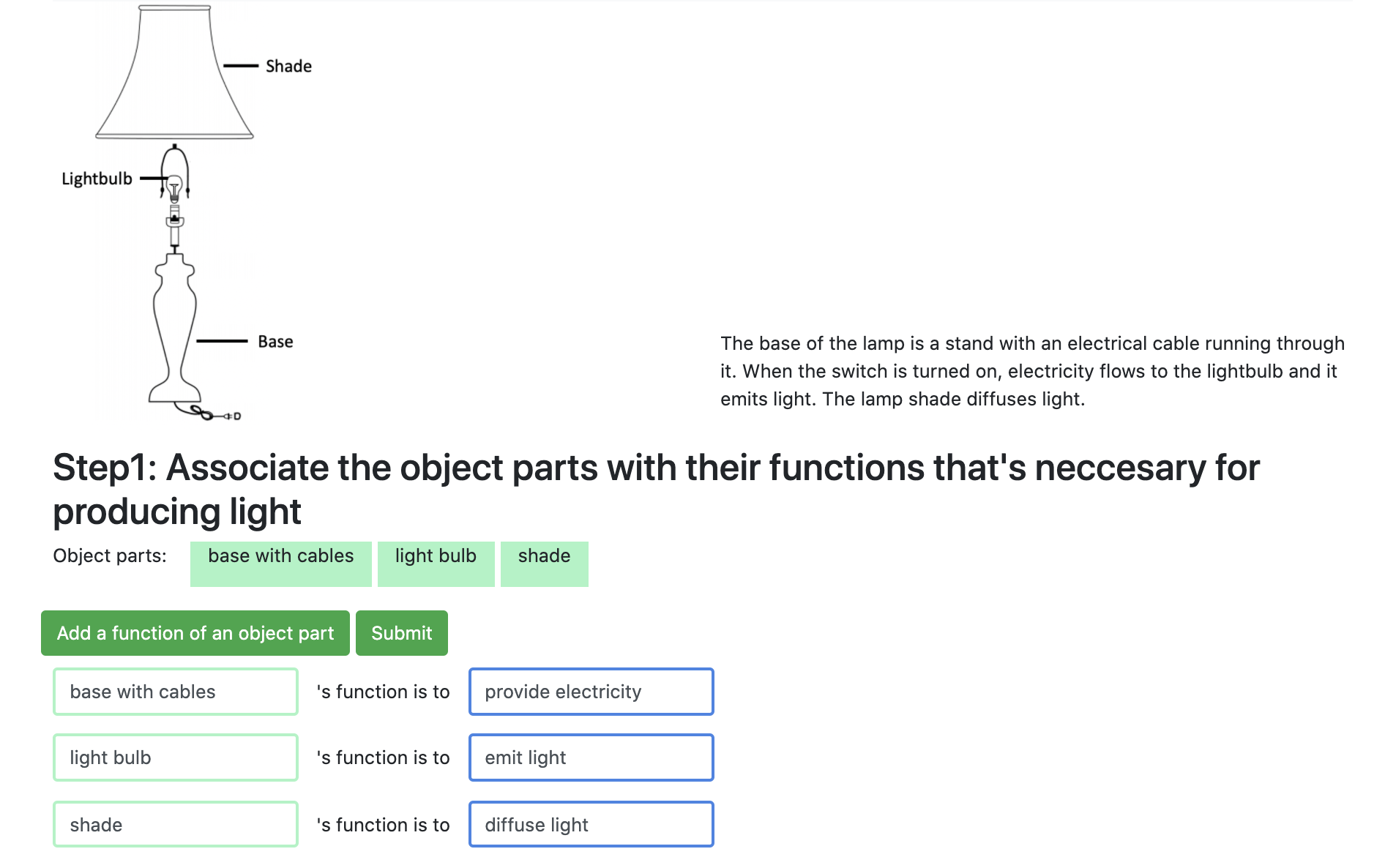}
    \caption{Step 1 user input example}
    \label{fig:step1}
\end{figure}

 Nine undergraduates attending Brown University  participated for course credit. They first read a short tutorial explaining that their task was to describe how an object produced a certain effect by constructing the causal model using our interface. The tutorial explained how the interface worked with an example object that was different than the light-producing objects. They were then directed to an experiment page that involved three steps. In the first step, they were presented with one of the four light-producing objects (order was counterbalanced across participants) with a short explanation of how the object worked, as well as a picture of the object with labeled parts. They identified the functions of object parts relevant to achieving the final effect - light. They did this by writing down function labels for object parts themselves as shown in Fig.\ref{fig:step1}. They were allowed to associate multiple functions with a single part, and multiple parts with a single function.\par

\begin{figure}[t]
    \centering
    \includegraphics[width=\linewidth]{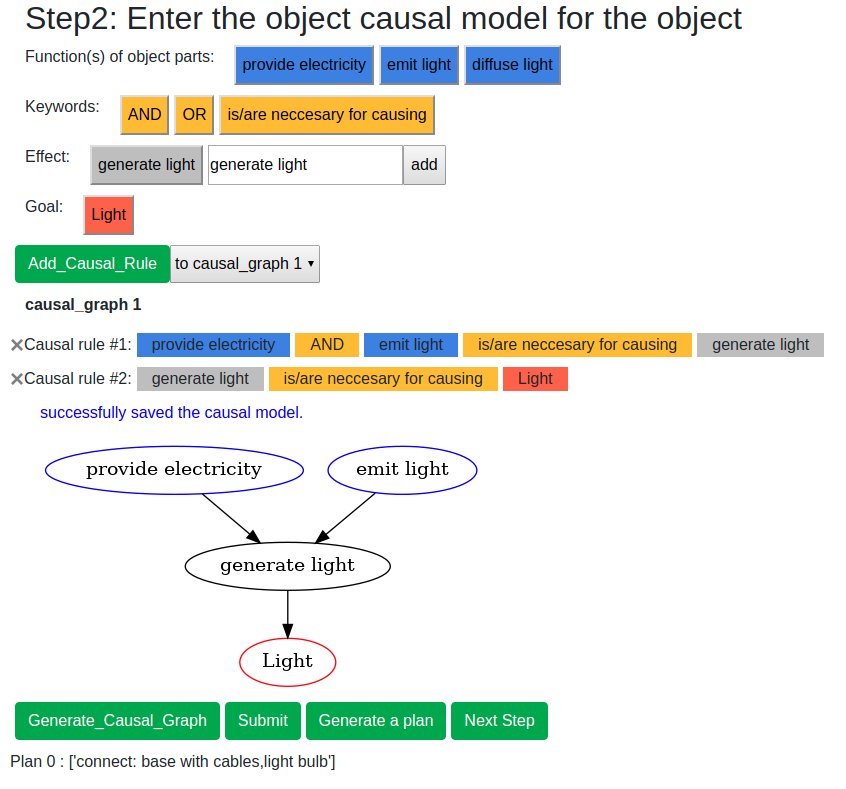}
    \caption{Causal model created by a user in Step 2 for the desk lamp. Functions of the object parts identified by the same user are shown in Fig.2}
    \label{fig:step2}
\vspace{-2mm}
\end{figure}

The second step involves describing the causal model of the object in terms of the functions of the object parts. The causal model is a collection of causal rules that are expressed using predefined keywords. Each rule establishes which function or multiple functions of object parts cause an effect. The final effect is always the goal (i.e., light), but participants were allowed to enter intermediate effects (that were different from functions of object parts identified in the first step) that can be used as causes in another causal rule. After they entered their causal model, they saw their model in graphical format and the interface generated a plan using their causal model and presented it to them in a text format describing which object parts are connected as shown in Fig.\ref{fig:step2}. If the generated plan was not satisfactory, participants could go back and update their causal models.\par

The presentation of the plan is intended to establish a dynamic communication channel between the human and the robot: the robot uses the causal model to generate a plan and the human sees how this causal model affects the robot's planning behavior, which can help the human understand what needs to be changed in the causal model if the plan is not satisfactory.
In the final step, we tested for near-generalization by presenting a new object from the same category as the training object and asked the participants to associate the functions of the new object parts with the functions they described for the previous object. Then, the planner tried to create a plan for the test object using the causal model that was created for the training object. Note that participants were not allowed to create a new causal model for the test object or to update their causal model of the training object. As a result, the planner could only succeed if the causal model of the training object was abstract enough to describe how the test object worked. Each participant went through this 3-step process once for each category.

\begin{figure}
    \centering
    \includegraphics[width=\linewidth]{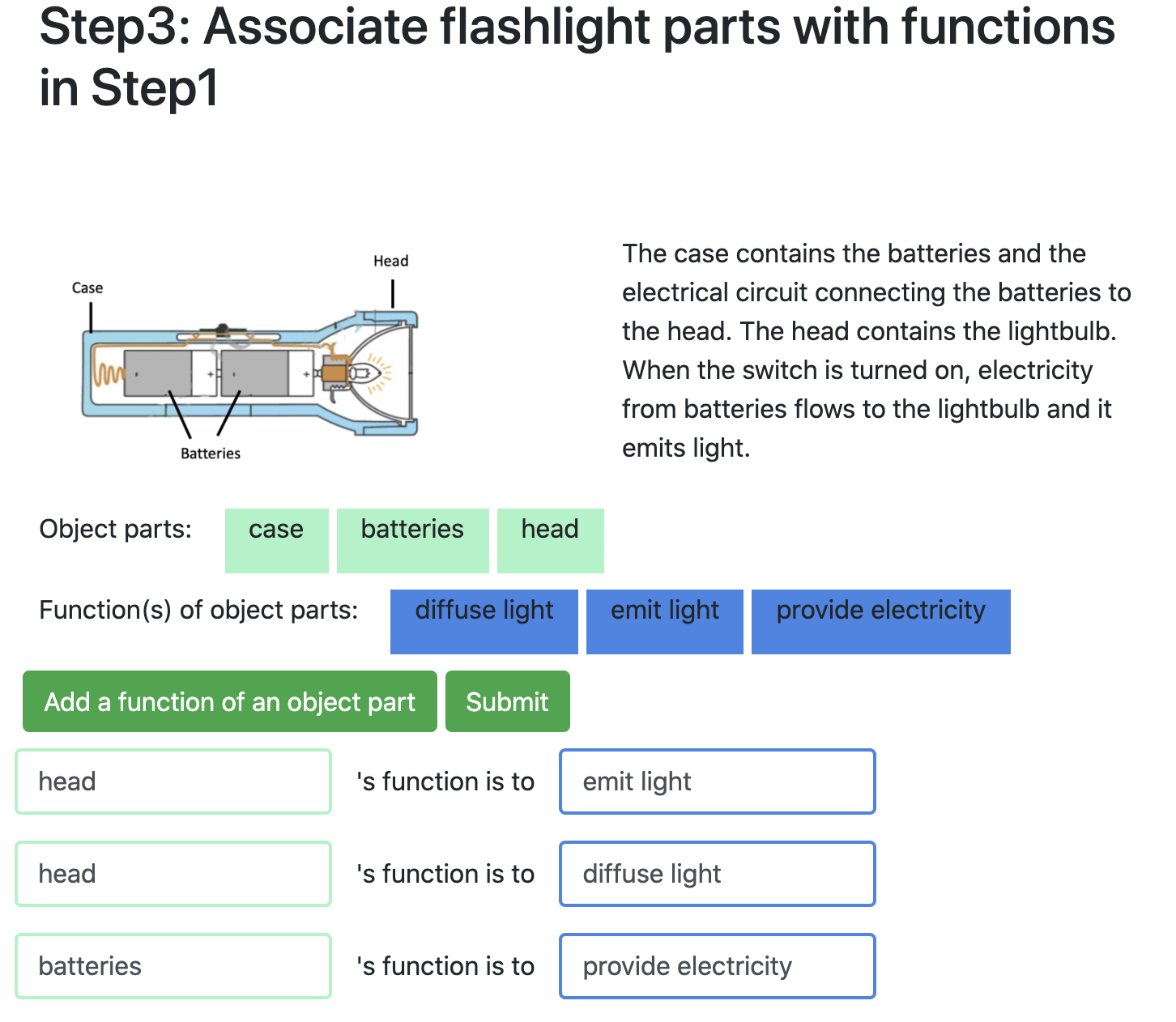}
    \caption{Step 3 user input example}
    \label{fig:step3}
\end{figure}

\subsection{Planner}
The planner is in charge of generating a sequence of actions. The actions that a planner generates are high-level action primitives such as, screw, insert, connect. The high-level action primitives will be processed to generate motor commands that can be performed by a simulator or a real robot. 
We choose to use Markov Decision Process (MDP) as the framework for planning. MDP captures the probability of state transition when an action is performed, which accounts for robot action uncertainty and object parts compatibility. The MDP problem is defined by $\{ S, A, T, R\}$, where $S$ represents the current assembly state (i.e., the current object parts assembled, their positions in the assembled structure), $A$ is the set of predefined actions, $T$ is the transition probability, and $R$ is the reward function at a given state.  The transition probability $T = P(S, A, S')$  can be estimated by the compatibility of the connection points between two object parts $O_1$, $O_2$. The connection compatibility can be pre-defined with binary values or can be estimated based on the geometric alignment of the connectors.

For the reward function $R$, we use the causal model $G$ created by the user to guide the reward for each state. The causal model generates a score representing whether a given state is able to achieve the final effect. Each node $V$ in the causal graph is a binary value. For a current assembled structure with object $O_1$ .. $O_n$, function nodes associated with each object parts will have value a of 1. The value of the final effect node, $V_{goal}$, is determined by a functional relationship $f$ of the values of its parent nodes and calculated recursively until the function nodes (i.e., root nodes) are reached:
 \begin{equation}
     V_{goal} = f(parent(V_{goal}))
 \end{equation}
The reward function is dependent on the value of $V_{goal}$. We assign positive reward if the goal is reached and negative reward if goal is not reached and there is no further applicable actions.

\section{Results}

The results of our preliminary experiment suggest that most people were able to use our interface to express reasonable causal models with minimal instructions that they received on the tutorial page. An example of a successful model created by a participant for the desk lamp (training) that transferred to the flashlight (test) is shown in Fig.\ref{fig:step2} and Fig.\ref{fig:step3}. 

This provides preliminary evidence supporting the claim that causal models created by people can achieve near-generalization. Moreover, it also shows that our interface can be used to communicate these causal models. In some cases, participants created valid causal models for the training object but their model failed to transfer to the test object. An example model that failed to transfer from desk lamp to flashlight is shown in Fig.\ref{fig:failed}. This model fails because of two reasons. First, the participant did not specify any of the flashlight parts with the function ``turn electricity into light" that was in their causal model, and as such, the flashlight parts were not enough to achieve the goal. Second, in step 3 the user added a new function ``hold things together" for the ``case" part of the flashlight that was not present in their earlier causal model. 

Some of the failures of transfer are likely due to participants misunderstanding the task. For instance, it is likely that the participant decided to include ``hold things together" as a function for the flashlight because they misunderstood the task to be describing the functions of all object parts instead of describing only the functions relevant to producing light. The same participant included ``diffuse light" function of the ``shade" of the desk lamp to be necessary for achieving the goal of producing light, which clearly cannot be true since one has to produce light before it can be diffused. In the next experiment, we will clarify subtle differences like these to elicit more appropriate causal models for our task. 

\begin{figure}[h!]
    \centering
    \includegraphics[width=\linewidth]{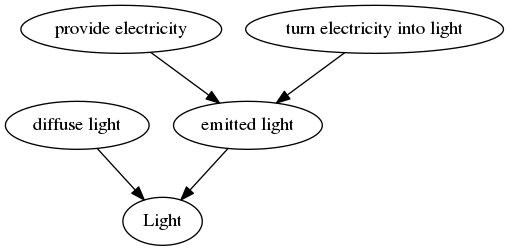}
\end{figure}

\begin{table}[h!]
    \centering
\begin{tabular}{|c|c|}\hline
desk lamp parts & functions \\ \hline
    shade & diffuse light \\
    light bulb & turn electricity into light \\
    base with cables & provide electricity \\\hline \hline
  flashlight parts & functions \\ \hline 
    case & hold things together \\
    head & diffuse light, provide electricity \\
    batteries & provide electricity \\\hline
  \end{tabular}
  \captionof{figure}{Example of a failed causal model generated by a user. Table on the top shows the step 1 input for the training object (desk lamp) and the table below shows the step 3 input for the test object (flashlight)}
    \label{fig:failed}
\end{table}

\vspace{-2mm}

 Finally, in a few trials participants generated invalid causal models. For instance, some users specified intermediate effect nodes in their causal models but did not specify their causes despite being instructed to do so in the tutorial. We will prevent the users from submitting such causal models and present them with an error message explaining what they need to change. 

\section{Far-generalization experiment}
For our next experiment, we will update our interface to address the ambiguities that led some participants to generate causal models that were invalid or failed to transfer, as discussed in the previous section. We will use the updated interface to test for both near-generalization and far-generalization. We will present two training objects to participants and ask them to generate a single causal model that describes how they produce light. Half of the participants will be assigned to a near-generalization condition where the two training objects they see will be from the same category (i.e., producing light by burning fuel or by using electricity). The other half will be in a far-generalization condition and they will see one object from each category. The test phase will be identical for both groups of participants: they will see a novel object from one category and will be asked to complete step 3 of the current experiment as described in the "Near-generalization experiment" section. Then, they will see a second test object from a different category than the first test object and will be asked to repeat step 3 with this object as well. We predict that presenting two objects from the same category will lead participants to generate causal models that will only generalize to the test object from the same category, so participants in the near-generalization condition are predicted to create models that will achieve near-generalization but not far-generalization. In contrast, presenting two objects from different categories to the participants in the far-generalization category should lead them to think in even more abstract terms, so we predict that their causal models will generalize to both test objects. The order of objects presented as training and test objects will be counterbalanced across participants.

\vspace{-2mm}
\section{Future Work}
\vspace{-2mm}
In addition to our next experiment, in the future, we are planning to use an online platform, such as Amazon Mechanical Turk, to recruit a high number of participants and test the generalizability of the causal models they generate. Inspired by recent developments in relation networks \cite{DBLP:journals/corr/abs-1711-06025} and graph neural networks \cite{DBLP:journals/corr/abs-1812-08434}, we aim to use a neural network to extract a single causal model from a large sample of user-generated causal models. We predict that a causal model learned from a large sample of users is more likely to be accurate compared to a model generated by an average user, given the evidence suggesting that collective decision making tends to outperform decisions of individuals \citep{surowiecki2005wisdom} even if the individuals are experts \citep{wolf2015collective}.

\bibliography{references.bib}
\end{document}